\documentclass[conference]{IEEEtran}
\IEEEoverridecommandlockouts
\usepackage{cite}
\usepackage{amsmath,amssymb,amsfonts}
\usepackage{algorithmic}
\usepackage{textcomp}
\usepackage{xcolor}
\usepackage{graphicx}
\usepackage{mathrsfs}
\usepackage{colortbl,booktabs}
\usepackage{multirow}
\def\BibTeX{{\rm B\kern-.05em{\sc i\kern-.025em b}\kern-.08em
    T\kern-.1667em\lower.7ex\hbox{E}\kern-.125emX}}
\begin{document}

\title{Linguistic-Enhanced Transformer with CTC Embedding for Speech Recognition
}


\author{\IEEEauthorblockN{Xulong Zhang, Jianzong Wang$^\ast$\thanks{$^\ast$Corresponding author: Jianzong Wang, jzwang@188.com.}, Ning Cheng, Mengyuan Zhao, Zhiyong Zhang, Jing Xiao}
\IEEEauthorblockA{\textit{Ping An Technology (Shenzhen) Co., Ltd.}}
}

\maketitle

\begin{abstract}
    The recent emergence of joint CTC-Attention model shows significant improvement in automatic speech recognition (ASR). The improvement largely lies in the modeling of linguistic information by decoder. The decoder joint-optimized with an acoustic encoder renders the language model from ground-truth sequences in an auto-regressive manner during training. However, the training corpus of the decoder is limited to the speech transcriptions, which is far less than the corpus needed to train an acceptable language model. This leads to poor robustness of decoder. To alleviate this problem, we propose linguistic-enhanced transformer, which introduces refined CTC information to decoder during training process, so that the decoder can be more robust. Our experiments on AISHELL-1 speech corpus show that the character error rate (CER) is relatively reduced by up to 7\%. We also find that in joint CTC-Attention ASR model, decoder is more sensitive to linguistic information than acoustic information.
\end{abstract}


\section{Introduction}
The adoption of end-to-end models greatly simplifies the training process of automatic speech recognition (ASR) system.
In such models, acoustic, pronunciation and language modeling components are jointly optimized in an unified system.
There's no need to train these components separately and then integrate them during decoding as traditional hybrid system \cite{bourlard1994connectionist}.
As the end-to-end model develops, it gradually exhibits superior performance than traditional hybrid system in both accuracy and real-time factor(RTF),
which makes it universal and a trend in the speech recognition research community.

The fundamental problem of end-to-end ASR is how to process input and output sequences of different lengths.
Two main approaches are proposed to handle this problem.
One of them is connectionist temporal classification (CTC) \cite{graves2006connectionist, graves2014towards}.
CTC introduces a special label ``blank", so that input sequence and output sequence can be aligned,
then the CTC loss can be effectively computed using forward-backward algorithm.
The other framework is attention based encoder-decoder (AED) model \cite{chorowski2015attention, 7472621}.
Chorowski \textit{et al.} introduce AED model into speech recognition \cite{chorowski2015attention}.
LAS \cite{7472621} produces character sequences without making any independence assumptions between the characters.
Furthermore, many optimizations of neural network structure or strategies,
such as Convolutional Neural Network (CNN) \cite{zhang2017very, yalta2019cnn}, Long-Short Term Memory (LSTM) \cite{zhang2017very}, Batch Normalization (BN) \cite{zhang2017very}, are conducted on both CTC and AED approaches.

The other framework is attention based encoder-decoder (AED) model \cite{gulcehre2015using, bahdanau2014neural, chorowski2015attention, 7472621, bahdanau2016end,zhang2022SUSing}.
This model firstly achieves great success on neural machine translation task \cite{gulcehre2015using, bahdanau2014neural,asru2021zhang}, then quickly expands to many other fields.
Chorowski \textit{et al.} introduce AED model into speech recognition \cite{chorowski2015attention}.
They get PER 17.6\% on the TIMIT phoneme recognition task, and solve the problem of accuracy decrease for long utterances by adding location-awareness to the attention mechanism.
LAS \cite{7472621} produces character sequences without making any independence assumptions between the characters.
\cite{bahdanau2016end} uses WFST to decode AED model with a word-level language model.

A major breakthrough is joint CTC-Attention model \cite{kim2017joint, watanabe2017hybrid,zhang2020research,bai2020listen, tian2020spike, higuchi2020mask, higuchi2021improved, fan2021cass,zhang2021singer,zhang2020unified, wang2021wnars} based on multi-task learning (MTL) framework proposed by Watanabe, kim \textit{et al.},
which fuses the advantages of the two approaches above.
CTC alignment is used as auxiliary information to assist the training of AED model.
Hence, the robustness and converge speed are both significantly improved.
This method gradually become standard framework of end-to-end ASR task.
Hori \textit{et al.} propose CTC, attention, RNN-LM joint decoding \cite{hori2017advances},
and introduce word-based RNN-LM \cite{hori2018end}, which further improve performance of end-to-end ASR.

Another important improvement is Transformer \cite{vaswani2017attention} proposed by Vaswani \textit{et al.}, which dispense the recurrent network in the AED model and based solely on attention mechanisms.
This significantly speedup the training, and save computing resources.
Dong \textit{et al.} apply Transformer on ASR task \cite{dong2018speech}, and get word error rate (WER) of 10.9\% on Wall Street Journal (WSJ) dataset.
\cite{gulati2020conformer} propose Conformer, a network structure combining CNN and Transformer, which further improves model performance.

Joint CTC-Attention model with multi-task learning framework produces 2 decodable branch: CTC and attention.
The attention branch often outperform CTC, because CTC requires conditional independence assumptions to obtain the label sequence probabilities.
However, attention branch has a drawback that word embedding is autoregressive so that the computational cost is very large, and it can not process decoding in parallel.
Many studies have been made to tackle this drawback.

In joint CTC-Attention model, the encoder is like an acoustic model, and the decoder is like a language model.
Despite the fusion and joint training of these two models bring quite a significant improvement,
the training corpus of the decoder (language model) is limited to the speech transcriptions,
which is far less than the corpus needed to train an acceptable language model.
We found that the involvement of CTC gives us the opportunity to generate more linguistic features to assist the training of decoder.
In this work, we propose linguistic-enhanced transformer, a simple training method, which introduces linguistic information refined from CTC encoder to the AED decoder.
In this way, CTC branch contributes more linguistic information to decoder during training.
On the other hand, compared with baseline approach, which loads linguistic target only from ground truth,
this method involves some "error" into linguistic target during the training process of AED decoder.
This reduces the inconsistency of training and decoding process.
Therefore, the decoder is trained to be more robust and brings performance improvement than baseline system.

\section{Methodology}
\label{sec:Methodology}
\subsection{Embedding Fusion}
\label{ssec:subhead}
As shown in Fig.1, the basic idea of embedding fusion (EF), is directly push CTC 1-best into attention decoder's word embedding layer just like the ground truth transcription.

\begin{figure}[htb]
\centering
\includegraphics[width=\linewidth]{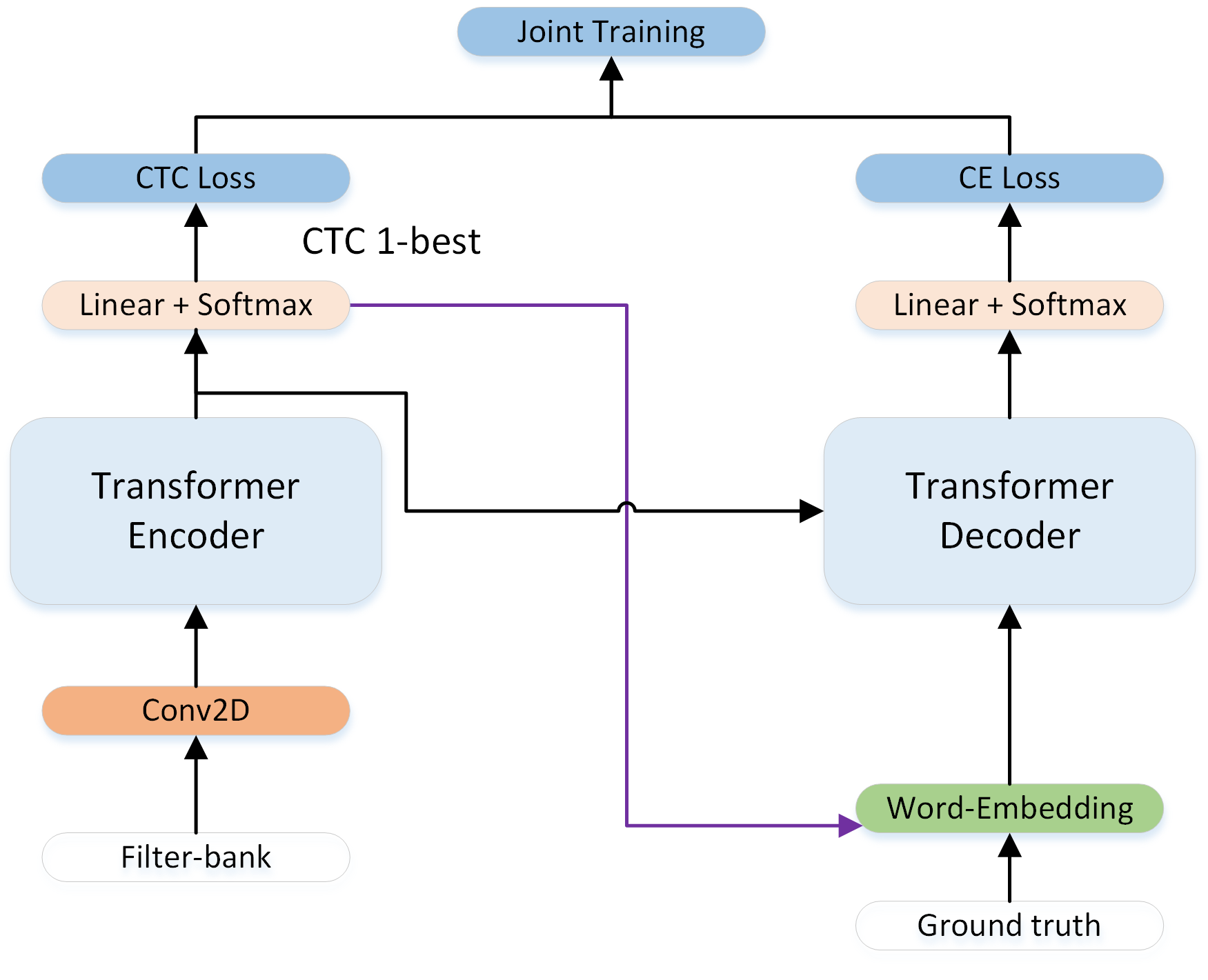}
\caption{Embedding Fusion (EF): CTC 1-best and ground truth share word-embedding of decoder.}
\end{figure}


We define the output of the acoustic encoder $h^{s}$ as:
\begin{equation}
h^{s} = E(x)
\end{equation}
where $E(\cdot)$ is the encoding function.
We process a Linear layer and a Softmax layer on $h^{s}$ to get CTC posterior,
and invoke greedy search to get CTC 1-best $W=(w_{1},\cdots,w_{L})$:
\begin{equation}
W = \mathcal{G}(Softmax(Linear(h^{s})))
\end{equation}
where $\mathcal{G}(\cdot)$ refers to greedy search algorithm.
Then we fuse the word-embeddings of ground truth transcription and CTC 1-best with linear combination:
\begin{equation}
\mathcal{E}_{fusion} = \alpha \cdot \mathcal{E}(W) + (1-\alpha) \cdot \mathcal{E}(y)
\end{equation}
where $\mathcal{E}(\cdot)$ refers to word-embedding function,
$W$ is CTC 1-best, $y$ is ground truth transcription, $\alpha$ is a tunable parameter to control the weights of the two text sources.
Finally, we feed $\mathcal{E}_{fusion}$ into decoder layers.

This may cause some issues.
For example, if the greedy CTC output have different length from ground truth transcription, they will have different shape after word-embedding,
so that they can not be combined with the equation above.
Especially when the training process was in the beginning, the model is not well trained, so that the CTC output will be disorder.

To solve this problem, we add some rules to our training process.
The basic idea is that if the CTC 1-best and ground truth transcription have identical lengths, we combine them using the equation above,
if the lengths of them are not equal, but relatively close, we use CTC 1-best as the decoder input,
if the lengths of them are too far apart, we feed ground truth to the decoder just like standard Transformer.
Obviously, we need to define some rules to judge what is ``relatively close" and what is ``far apart".
We design 2 types of threshold.
First is absolute threshold.
Let $T_{l}$ denotes maximum allowance length difference between the ground truth transcription and CTC 1-best,
If $|L_{ctc} - L_{groundtruth}| \leq T_{l}$, we think their lengths are relatively close.
The other is relative threshold $T_{r}$, which consider the length differences of ground truth sentences.
$|L_{ctc} - L_{groundtruth}| / L_{groundtruth} \leq T{r}$ indicates that lengths are relatively close.

\subsection{Aligned Embedding Fusion}
Inspired by \cite{higuchi2021improved},
we find that even though we add some rules to prevent the poor CTC output from being fed to AED decoder, there are still a defect in EF method.
The problem is that CTC 1-best and ground truth transcription are not aligned.
So if we set them to be the input and output of AED decoder respectively,
AED decoder may see wrong training samples.
If CTC 1-best has the same length as ground truth, and has only substitution errors compared with ground truth, this problem will not show up.
In this situation, AED decoder will be trained to be a ``corrector".
But when the length of CTC 1-best and ground truth differ, or they have the same lengths but their text are dislocated,
actually when CTC 1-best has insertion errors or deletion errors compared with ground truth,
this problem will become obvious.

For example, ground truth is $y=\{A, B, C, A\}$, CTC greedy output is $W=\{A, C, A\}$.
In step 2 of decoder training, in normal case, we feed $\{sos, A, B\}$ into decoder, and wish decoder to predict $\{C\}$ as the next token.
If we use CTC 1-best as input, we put $\{sos, A, C\}$ and wish it to predict $\{C\}$. This is a wrong training sample.

To alleviate this shortcoming, we propose aligned embedding fusion (AEF),
which align the CTC 1-best and ground truth with edit-distance algorithm and insert special symbol ``blank" into the specific position.
\begin{equation}
(y_{align}, W_{align}) = \mathcal{A}_{text}(y, W)
\end{equation}
where $\mathcal{A}_{text}(\cdot)$ represents text alignment with edit distance algorithm,
$y_{align}$ and $W_{align}$ denote ground truth and CTC 1-best after text alignment respectively.
In the example above, the CTC greedy above will change to $W_{align} = \{A, blank, C, A\}$, and ground truth is unchanged $y_{align} = \{A, B, C, A \}$.
Then we can feed them into decoder just like EF.

\subsection{N-best Embedding}

\begin{figure}[htb]
\centering
\includegraphics[width=\linewidth]{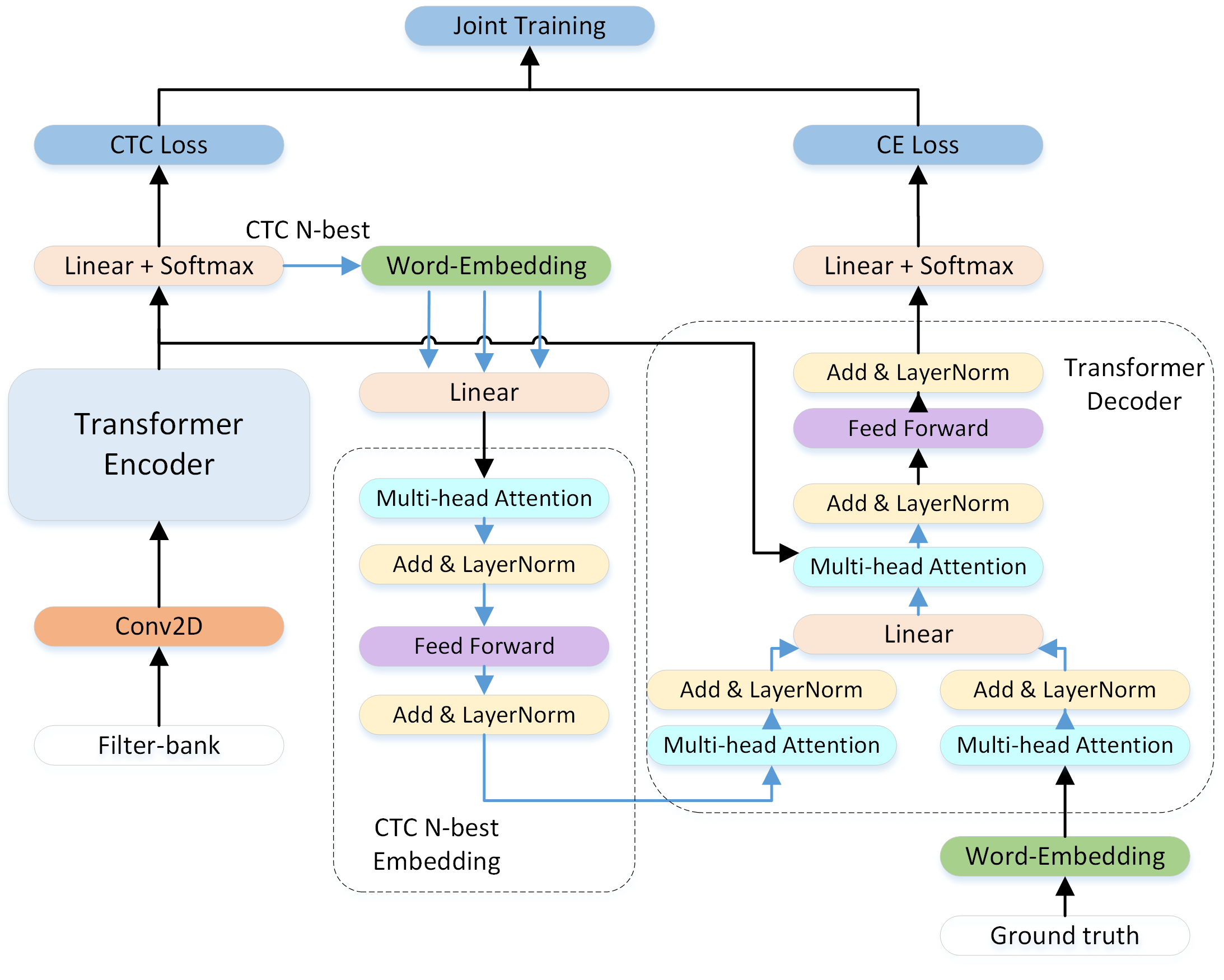}
\caption{N-best Embedding (NE): Extract CTC N-best information by attention machanism, and feed it into Transformer decoder.}
\end{figure}

Since CTC 1-best has been introduced,
we try to feed CTC N-best information to attention decoder instead of CTC 1-best.
Then we proposed ``N-best embedding" (NE).
Fig.2 shows the architecture of NE model.

Firstly, we adopt beam-search instead of greedy-search to get CTC N-best.
\begin{equation}
\mathcal{N} = \mathcal{B}(Softmax(Linear(h^{s})))
\end{equation}
where $\mathcal{N} = \{N_{1}, N_{2} .... N_{n}\}$ refers to CTC N-best, $\mathcal{B}$ refers to beam search algorithm.
$n$ is controlled by the ``beam" of beam-search algorithm.
Secondly, we use the same word-embedding layer as decoder, to project each N-best text to a ${d_{model}}$ dimensional vector, then concatenate them together.
Next, we use a linear layer to project this $n * {d_{model}}$ dimensional vector back to a ${d_{model}}$ vector.
\begin{equation}
Input_{NE} = Linear(Concat(\mathcal{E}(\mathcal{N})))
\end{equation}
$Input_{NE}$ will be fed into the following ``CTC N-best Embedding" module.

``CTC N-best Embedding" module consists of several identical self-attention layers, and each layer is similar to encoder layer.

Finally, we modify the decoder.
We add an additional attention sub-layer in every decoder layer.
This sub-layer locates side by side with the self-attention sub-layer, and takes responsibility for accepting information from CTC N-best Embedding module.
The output of this sub-layer will be concatenated with the output of self-attention sub-layer,
then projected back to ${d_{model}}$ by a linear layer, so it can be sent to the subsequent ``encoder-decoder attention" sub-layer.

\section{Experiments}
\label{sec:Experiments}

\subsection{Data}
All the experiments are conducted on AISHELL-1 speech corpus.
AISHELL-1 dataset consist of about 170 hours Mandarin audio data of 400 speakers with corresponding text transcriptions.
The size of text transcriptions is about 6M.
Our model use 4233 distinct labels: 4231 characters, blank, and sos/eos tokens.
We extracted 80-channel log-mel energy from a 25ms window, and shift the window every 10ms to get our acoustic feature vectors.
Table \ref{tab:1.0} shows the sentence length (in character) distribution in AISHELL-1 dataset.

\begin{table}
\centering
\caption{Sentence length (in character) distribution in AISHELL-1 dataset.}
\begin{tabular}{|c|c|c|c|c|c|c|}
\toprule
& \multicolumn{6}{c|}{Sentence percentage} \\
\hline
Subset & 1-5 & 6-9 & 10-14 & 15-19 & 20-24 & $\ge 25$ \\
\hline
train & 0.16\% & 12.02\% & 42.85\% & 29.64\% & 14.94\% & 0.39\% \\
dev & 0.10\% & 12.52\% & 43.07\% & 29.22\% & 14.85\% & 0.24\% \\
test & 0.04\% & 11.16\% & 41.57\% & 30.82\% & 16.08\% & 0.32\% \\
\bottomrule
\end{tabular}
\label{tab:1.0}
\end{table}

\subsection{Model Setup}
In this work, Transformer \cite{vaswani2017attention} based on joint CTC-Attention multi-task learning (MTL) framework \cite{kim2017joint, watanabe2017hybrid} are adopted as our baseline.
In our Transformer model, encoder contains a 2D-convolution input layer and 12 encoder layers.
Decoder has 6 decoder layers.
CTC N-best embedding module in NE method contains 2 layers.
The dimension of self-attention sub-layer in encoder layer and decoder layer are both 256,
and the dimension of linear output layer is 1024.
The heads number in multi-head attention layer is 4.
Training mini-batch are set to 32, weight of CTC-loss is 0.3.
Beam-search algorithm is adopted during testing and beam value is set to 10.
No external language models are introduced.
All the experiments are implemented using ESPnet toolkit.

\subsection{Results}

Not surprising that EF model brings an improvement compared with the baseline Transformer.
AEF model looks better, but the increase is not large compared with EF model.
To find out the reason, we counted the number of blanks inserted during the training process of AEF (shown in Fig.3).
In the beginning of training, the number is so high,
but as the training goes on, it significantly decreased.
It seems that after several epochs of iterations, the CTC branch often gives quite accurate length prediction,
so that there is only a small part of training samples that really need to be aligned.

N-best embedding model (NE) shows insignificant improvement when $n=1$.
As the $n$ increases, NE model get better and gradually outperform AEF model.
When $n=5$, the performance of NE model goes stable.

\begin{table}
\centering
\caption{The character error rate (CER) of baseline and proposed approaches on AISHELL-1 dataset.}
\begin{tabular}{c|c|c|c}
\toprule
Training Method & Model config & Decode method & Dev/Test\\
\hline
 Transformer & - & \multirow{7}{*}{Attention Decoder} & 6.7/7.7\\
\hline
EF & $T_{l}$=2 &  & 6.6/7.5\\
EF & $T_{r}$=0.15 &  & 6.6/7.5\\
AEF & $T_{l}$=2 &  & \textbf{6.5}/7.5\\
AEF & $T_{r}$=0.15 &  & \textbf{6.5}/7.5\\
NE & $n$=1 &  &  6.8/7.8\\
NE & $n$=5 &  &  \textbf{6.5}/\textbf{7.4}\\
\hline
Transformer & - & \multirow{7}{*}{CTC + Rescore} & 7.2/8.3\\
EF & $T_{l}$=2 &  & 6.9/7.9\\
EF & $T_{r}$=0.15 &  & 6.9/8.0\\
AEF & $T_{l}$=2 &  & \textbf{6.7}/7.8\\
AEF & $T_{r}$=0.15 &  & 6.8/7.8\\
NE & $n$=1 &  &  7.0/8.1\\
NE & $n$=5 &  &  \textbf{6.7}/\textbf{7.7}\\
\bottomrule
\end{tabular}
\label{tab:1.1}
\end{table}

Table \ref{tab:1.1} summaries the experimental results on AISHELL-1 corpus.
We test with 2 different decoding methods,
Attention decoding and CTC + Attention rescore.
We can observe that, by introducing CTC information, all proposed methods outperform baseline Transformer.
This indicates that with more linguistic information introduced, decoder was trained to be more robust.


\begin{figure}[htb]
\centering
\includegraphics[width=\linewidth]{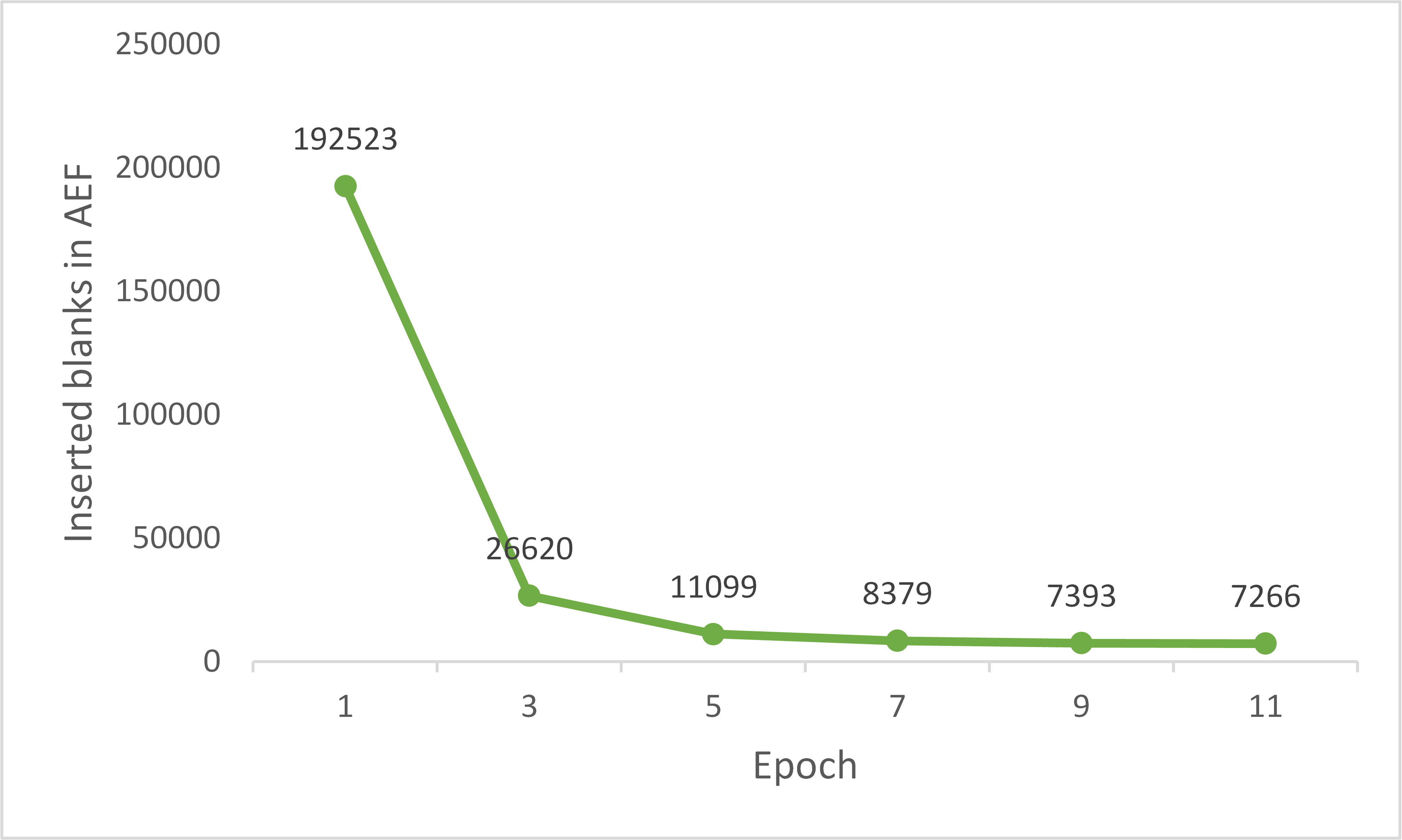}
\caption{Number of blanks inserted in AEF.}
\end{figure}

\begin{figure}[htb]
\includegraphics[width=0.32\linewidth]{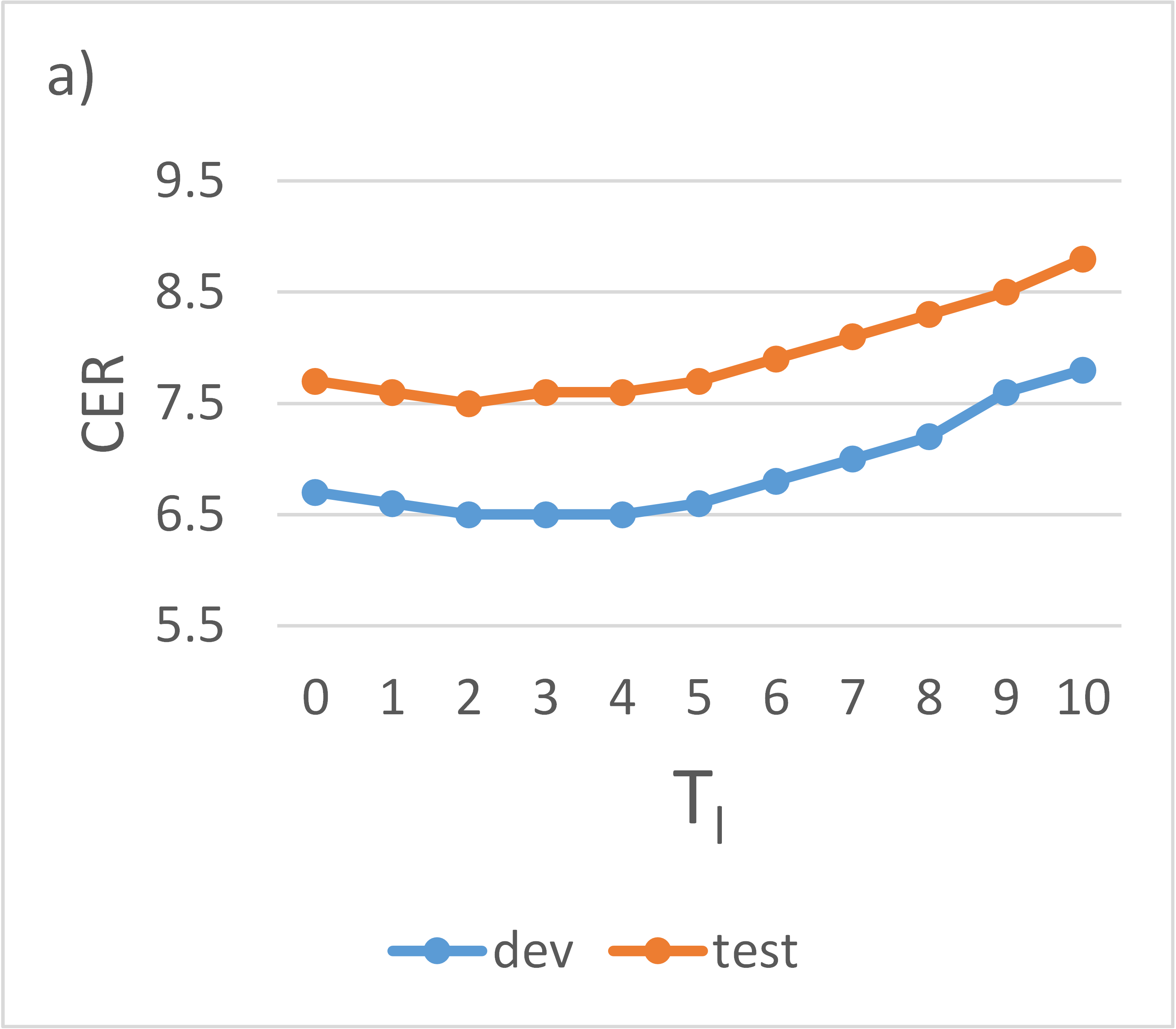}
\includegraphics[width=0.32\linewidth]{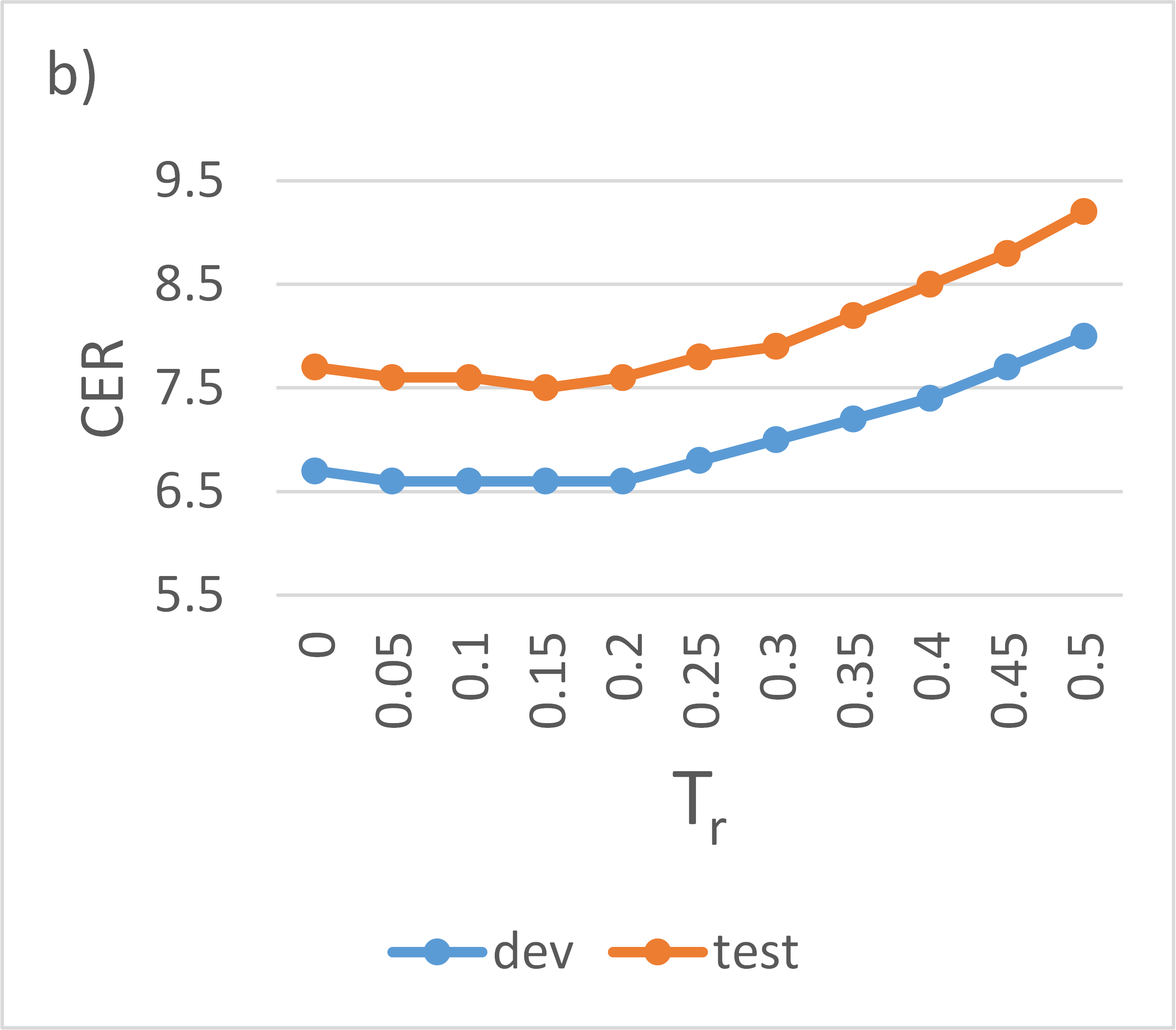}
\includegraphics[width=0.32\linewidth]{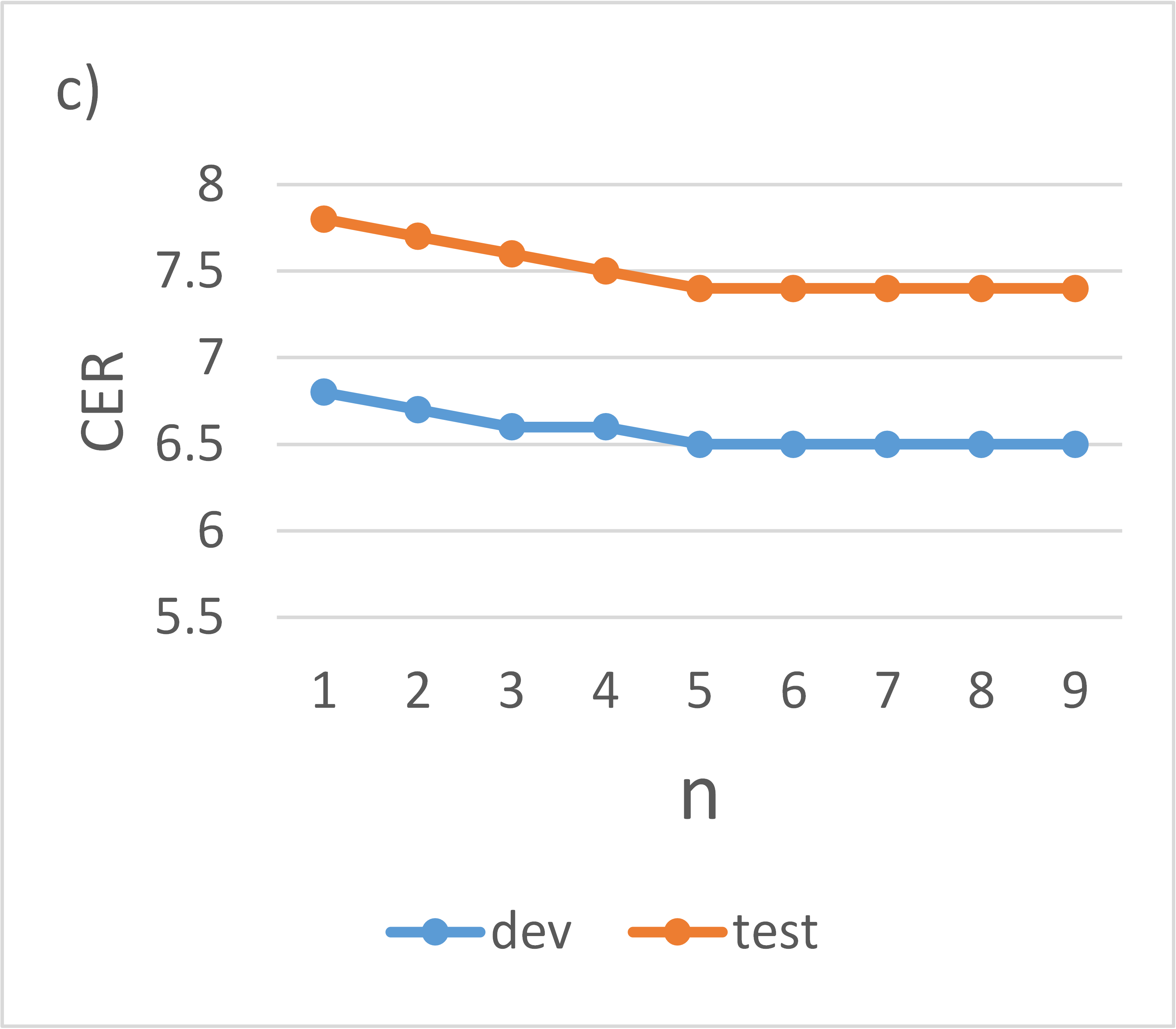}
\caption{a) In Aligned Embedding Fusion (AEF) model, $T_{l} = 2$ shows the best performance.
b) In AEF model, $T_{r} = 0.15$ shows the best performance.
c) In N-best Embedding (NE), $n = 5$ shows the best performance.}
\end{figure}

\subsubsection{Effect of $T_{l}$ and $T_{r}$}

We can observe that Fig.4a and Fig.4b have similar trend.
As the threshold goes up, the performance improved, but after reaching a critical point, the performance start to decrease.
The critical point of $T_{l}$ is 2, and for $T_{r}$, it is about 0.15.
It is easy to comprehend this.
Larger threshold means the attention decoder have more chance to see training samples with different lengths, which lead to stronger robustness.
But this may also introduce more errors.
Smaller threshold brings less errors along with less diversity of training samples.
They reach a balance at $T_{l} = 2$ and $T_{r} = 0.15$.
From the result of current experiment, 2 types of threshold give similar performance.
But we think that the optimal value of $T_{l}$ may be more sensitive to the data set and the modeling unit.
$T_{r}$ may be a better choice.


\subsubsection{Effect of $n$ in N-best embedding}

Fig.4c shows the effect of $n$ in NE model.
We can see that when $n$ is small, the performance of NE model increases with the increase of $n$.
When $n$ is relatively large, the performance of NE model no longer increases, but tends to stabilize.
The critical point of $n$ is about 5.
Larger $n$ means more redundant information, but it also makes the model more complex and slower to train due to beam-search.

\subsubsection{Effect of Pre-training}

\begin{table}
\centering
\caption{Performance comparison of Transformer, AEF and NE with Pre-training enabled.}
\begin{tabular}{c|c|c}
\toprule
Training Method & Pre-training module & Dev/Test\\
\hline
\multirow{3}{*}{Transformer} & None & 6.7/7.7\\
 & Encoder & 6.6/7.6\\
 & Encoder and Decoder & 6.6/7.6\\
\hline
\multirow{3}{*}{AEF} & None & 6.5/7.5\\
 & Encoder & \textbf{6.3}/\textbf{7.2}\\
 & Encoder and Decoder & \textbf{6.3}/7.3\\
 \hline
\multirow{2}{*}{NE} & None & 6.5/7.6\\
 & Encoder & 6.4/7.4\\
\bottomrule
\end{tabular}
\label{tab:3}
\end{table}

To a considerable extent, the correctness of CTC 1-best or N-best affects the effectiveness of our method.
So we test AEF and NE with pre-training enabled,
which means we initialize the encoder (or decoder) with a pre-trained standard Transformer model's encoder (or decoder).
As shown in Table \ref{tab:3}, with pre-training enabled, the error rates of AEF and NE are greater decreased than baseline Transformer.
Pre-training looks more effective for AEF model, especially encoder pre-training.
For NE model, encoder pre-training also brings an improvement but not much.

\subsection{Analysis}
We are curious that why our methods bring performance improvements.
In the baseline Transformer model, all the encoder information has already fed into decoder layers by attention mechanism.
All the CTC information including 1-best, N-best, is already contained in the information above.
So why does CTC outputs still benefit decoder?

One possible explanation is that, in the standard framework, encoder information before linear and softmax layer is ``implicit".
In EF and AEF, we use text level CTC 1-best, and in NE we use information extracted from text level CTC N-best.
No matter what kind of information, it is more ``explicit" than baseline.

The other difference is the position where the decoder accepts the information from encoder.
In baseline framework, decoder accept encoder information in the middle layers of decoders as acoustic features.
While in EF, AEF, decoder accept CTC 1-best around the word embedding module as linguistic features.
We know that the decoder in joint CTC-Attention model is more like a language model.
To some extent, we can say that decoder is more sensitive to linguistic features than acoustic features.
As a ``language model", the decoder prefer more explicit text level information to train itself.

These can explain why NE method doesn't perform better than AEF when $n=1$.
In NE, ``N-best embedding module" take responsibility to extract information from text level CTC N-best.
In addition, NE model doesn't accept CTC information from word embedding module as linguistic features,
Instead, it accepts information in the decoder layers.
These make the information ``implicit", then N-best's advantages not reflected.
Of course as $n$ goes up, NE eventually exceeded AEF.

\section{Conclusion}
\label{sec:Conclusion}
In this paper, we propose linguistic-enhanced transformer, which introduce CTC outputs into the AED decoder during training process,
to provide more text corpus to AED decoder, to make the decoder a better ``language model".
Specifically, EF and AEF introduce 1-best and NE introduces N-best to AED decoder.
Results on the AISHELL-1 corpus demonstrate that with text level CTC output embedded,
AED decoder learns more linguistic information, and becomes more robust.
Further more, we found that decoder in joint CTC-Attention framework is more sensitive to linguistic-level input than acoustic-level input.
Explicit information is more useful than implicit one for AED decoder training.
Therefore, NE method does not fully utilize the advantages of N-best compared with EF and AEF.
But these are instructive for us on how to train a better AED decoder for ASR task.


\section{Acknowledgement}
This paper is supported by the Key Research and Development Program of Guangdong Province under grant No.2021B0101400003. Corresponding author is Jianzong Wang from Ping An Technology (Shenzhen) Co., Ltd (jzwang@188.com).

\bibliographystyle{IEEEtran}
\bibliography{mybib}
\end{document}